\pdfoutput=1
\documentclass[10pt, a4paper]{article}
\usepackage{lrec2006}
\usepackage{amsmath}
\makeatletter
\newcommand{\@BIBLABEL}{\@emptybiblabel}
\newcommand{\@emptybiblabel}[1]{}
\makeatother
\usepackage{fancyhdr} 
\usepackage{hyperref}

\title{Annotating Cognates and Etymological Origin in Turkic Languages}

\name{Benjamin S. Mericli$^{\ast}$, Michael Bloodgood$^{\dagger}$} 

\address{$^{\ast}$University of Maryland, College Park, MD bmericli@umd.edu \\
         $^{\dagger}$University of Maryland, College Park, MD meb@umd.edu \\     
}         

\abstract{Turkic languages exhibit extensive and diverse etymological relationships among lexical items. 
These relationships make the Turkic languages promising for exploring automated translation lexicon induction by 
leveraging cognate and other etymological information. However, due to the extent and diversity of the types of relationships between 
words, it is not clear how to annotate such information. 
In this paper, we present a methodology for annotating cognates and etymological origin 
in Turkic languages. 
Our method strives to balance the amount of research effort the annotator expends with the utility of the annotations for 
supporting research on improving automated translation lexicon induction.}

\begin{document}

\thispagestyle{fancy}

\maketitleabstract

\section{Introduction}
Automated translation lexicon induction has been investigated in the literature and shown to be feasible for various language families
and subgroups, such
as the Romance languages and the Slavic languages \cite{mann2001,schafer2002}. Although there have been some studies investigating using 
Swadesh lists of words to identify Turkic language groups and loanword candidates \cite{vanderArk2007}, we are not aware of 
any work yet on automated translation lexicon induction for the Turkic languages.  

However, the Turkic languages are well suited to exploring such technology since they exhibit many diverse 
lexical relationships both within family and to languages outside of the family through loanwords. 
For the Turkic languages, it is prudent to leverage both cognate information and other etymological information
when automating translation lexicon induction. 
However, we are not aware of any corpora for the Turkic languages that have been annotated for this 
information in a suitable way to support automatic translation lexicon induction. 
Moreover, performing the annotation is not straightforward because of the 
range of relationships that exist. In this paper, we lay out a methodology for performing this annotation that is intended to 
balance the amount of effort expended by the annotators with the utility of the annotations for supporting computational 
linguistics research. 

\section{Main Annotation System}

We obtained the dictionary of the Turkic languages \cite{oztopcu1996}. One section of this dictionary contains 1996 English glosses and
for each English gloss a corresponding translation in the following eight Turkic languages: Azerbaijani, Kazakh, Kyrgyz, Tatar, Turkish, Turkmen,
Uyghur, and Uzbek. Table~\ref{t:aliveUnannotated} shows an example for the English gloss `alive.' 
When a language has an official Latin script, that script is used. Otherwise, the dictionary's transliteration is shown in parentheses. 
Our annotation system is to annotate 
each Turkic word with a two-character code. The first character will be a number indicating which words are cognate with each other and the
second character will indicate etymological information. Subsection~\ref{cognates} discusses how to define and annotate cognates and
subsection~\ref{etymology} discusses how to define and annotate etymological information.

\begin{table*}[hp] 
\begin{center}
\begin{tabular}{|l|l|l|l|l|l|l|l|} 
      \hline
      Azerbaijani & Kazakh & Kyrgyz & Tatar & Turkish & Turkmen & Uyghur & Uzbek \\ \hline
      canl\i & (tiri) & (t\"{u}r\"{u}\"{u}) & (janl\i) & canl\i & diri & (tirik) & tirik \\ \hline
\end{tabular}
\caption{Example entry from the eight-way dictionary for the English gloss `alive.'}
\label{t:aliveUnannotated}
 \end{center}
\end{table*}

\subsection{Cognates} \label{cognates}

According to the Oxford English Dictionary Online\footnote{http://www.oed.com/view/Entry/35870?redirectedFrom=cognate} accessed on February
2, 2012, `cognate' is defined as: ``...Of words: Coming naturally from the same root, or
representing the same original word, with differences due to subsequent separate phonetic development; thus, English \emph{five}, Latin
\emph{quinque}, Greek $\pi\varepsilon\nu\tau\varepsilon$, are cognate words, representing a primitive \emph{*penke}.'' 
As this definition shows, shared genetic origin is key to the notion of cognateness. A word is only considered cognate with another if both words proceed from
the same ancestor. Nonetheless, in line with the conventions of previous research in computational linguistics, we set a broader definition. 
We use the word `cognate' to denote, as in \cite{kondrak2001}: ``...words in different languages that are similar in form and meaning, without
making a distinction between borrowed and genetically related words; for example, English `sprint' and the Japanese borrowing `supurinto' are
considered cognate, even though these two languages are unrelated.''
These broader criteria are motivated by the ways scientists develop and use cognate identification algorithms in natural language processing (NLP)
systems. For cross-lingual applications, the advantage of such technology is the ability to identify words for which similarity in
meaning can be accurately inferred from similarity in form; it does not matter if the similarity in form is from strict genetic
relationship or later borrowing.

However, not every pair of apparently similar words will be annotated as cognate. For them to be considered cognates, the differences in form
between them must meet a threshold of consistency within the data. We will explain the definitions and rules for the annotators to follow in
order to establish such a threshold.

First, we elaborate on how our notion of cognate differs from that of strict genetic relation. 
At a high level, there are two cases to consider: A) where the words involved are native Turkic words, and B) where the words involved are
shared loanwords from non-Turkic languages. 
Within case A, there are two cases to consider: (A1) genetic cognates; and (A2) intra-family loans.
Table~\ref{t:A1Example} shows an example of case A1. This example shows the English gloss `one' for all eight Turkic languages,
descended from the same postulated form, \emph{*bir}, in Proto-Turkic \cite{ronatas2006}. 
Case A1 is the strict definition of `cognate,' and these
are to be annotated as cognate.

\begin{table*}[hp] 
\begin{center}
\begin{tabular}{|l|l|l|l|l|l|l|l|} 
      \hline
      Azerbaijani & Kazakh & Kyrgyz & Tatar & Turkish & Turkmen & Uyghur & Uzbek \\ \hline
      bir & (bir) & (bir) & (ber) & bir & bir & (bir) & bir \\ \hline
\end{tabular}
\caption{Example of case A1: genetic cognates. The English gloss is `one.'}
\label{t:A1Example}
 \end{center}
\end{table*}

Case A2 is for intra-family loans, i.e., a word of ultimately Turkic origin borrowed by one Turkic language from another Turkic language. These cases, contrary to the
strict definition, are to be marked as cognate in our system. An example is the modern Turkish neologism \emph{alma\c{s}} `alternation,
permutation', incorporated from the Kyrgyz (\emph{alma\c{s}}) `change' \cite{TDK1942}.  While rare, it is used today in Turkish 
scholarly literature to describe
concepts in areas such as mathematics and botany. Processing genetic cognates (case A1) and intra-family loans (case A2) differently 
would have little
impact on the success of a cross-dictionary lookup system. In fact, accounting for the difference might limit the efficacy of such a 
system.
Also, the time depth of intra-Turkic borrowings may be centuries or mere decades. The more distant the borrowing the more difficult 
it will
be for annotators to distinguish between cases A1 and A2. Hence, instances of case A2 are to be annotated as cognate in our
system.\footnote{For similar reasons, false cognates may be annotated as cognate if the annotator does not have readily available knowledge
indicating that they are false cognates. Although this is a potential limitation of our system, it is not clear how to distinguish false 
cognates from true cognates without significant additional annotation expense.} 

Case B is for situations of shared loanwords, where the source of the words is ultimately non-Turkic. 
There are three subcases: (B1) loanwords borrowed from the same non-Turkic language; (B2) loanwords borrowed from different non-Turkic
languages, but of the same ultimate origin; and (B3) loanwords of non-Turkic origin borrowed via another Turkic language. 

Table~\ref{t:B1Example} shows an example of case B1, the word `book,' borrowed from Arabic in all eight Turkic languages. 
Table~\ref{t:B2Example} shows an example of case B2, the word `ballet,' borrowed from Russian in all cases except Turkish, where it was
borrowed directly from the French. 
Table~\ref{t:B3Example} shows an example of case B3: the word `benefit' in Kyrgyz was borrowed most likely through Uzbek or Chaghatay
\cite{kirchner2006}, but the Uzbek word was borrowed from Persian, and ultimately from Arabic.
It is difficult and time-consuming for annotators to make these fine-grained distinctions. And again, for computational processing,
such distinctions are not expected to be helpful. Hence, all of cases B1, B2, and B3 are to be annotated as cognate in our system. 

\begin{table*}[hp] 
\begin{center}
\begin{tabular}{|l|l|l|l|l|l|l|l|} 
      \hline
      Azerbaijani & Kazakh & Kyrgyz & Tatar & Turkish & Turkmen & Uyghur & Uzbek \\ \hline
      kitab & (kitap) & (kitep) & (kitap) & kitap & kitap & (kitab) & kitob \\ \hline
\end{tabular}
\caption{Example of case B1: loanwords borrowed from the same non-Turkic language. The English gloss is `book.'}
\label{t:B1Example}
 \end{center}
\end{table*}
 
\begin{table*}[hp] 
\begin{center}
\begin{tabular}{|l|l|l|l|l|l|l|l|} 
      \hline
      Azerbaijani & Kazakh & Kyrgyz & Tatar & Turkish & Turkmen & Uyghur & Uzbek \\ \hline
      balet & (balet) & (balet) & (balet) & bale & balet & (balet) & balet \\ \hline
\end{tabular}
\caption{Example of case B2: loanwords borrowed from different non-Turkic languages, but of the same ultimate origin. The English gloss is `ballet.'}
\label{t:B2Example}
 \end{center}
\end{table*}

\begin{table*}[hp] 
\begin{center}
\begin{tabular}{|l|l|l|l|l|l|l|l|} 
      \hline
      Azerbaijani & Kazakh & Kyrgyz & Tatar & Turkish & Turkmen & Uyghur & Uzbek \\ \hline
      fayda & (payda) & (payda) & (fayda) & fayda & pe\'{y}da & (payda) & foyda \\ \hline
\end{tabular}
\caption{Example of case B3: loanwords of non-Turkic origin borrowed via another Turkic language. The English gloss is `benefit.'}
\label{t:B3Example}
 \end{center}
\end{table*}

Recall that all our annotations are two-character codes; the first character is a number from one to eight indicating what words are cognate
with each other. 
Table~\ref{t:cognatesExample} shows the first character of the annotations for the example from Table~\ref{t:aliveUnannotated}.
The words marked with 1 are cognate with each other and the words marked 2 are cognate with each other. 

\begin{table*}[hp] 
\begin{center}
\begin{tabular}{|l|l|l|l|l|l|l|l|} 
      \hline
      Azerbaijani & Kazakh & Kyrgyz & Tatar & Turkish & Turkmen & Uyghur & Uzbek \\ \hline
      canl\i & (tiri) & (t\"{u}r\"{u}\"{u}) & (janl\i) & canl\i & diri & (tirik) & tirik \\ \hline
      1 & 2 & 2 & 1 & 1 & 2 & 2 & 2 \\ \hline
\end{tabular}
\caption{Example with cognates annotated.}
\label{t:cognatesExample}
 \end{center}
\end{table*}

\subsection{Etymology} \label{etymology}

The second character in a word's annotation indicates a conjecture about etymological origin, e.g., T for Turkic. 
The decision to annotate word origin is motivated by its value for facilitating the development of technology for cross-language lookup of
unknown forms. We therefore take a practical approach, balancing the value of the annotation for this purpose with the amount of effort
required to perform the annotation. We have created the following code for annotating etymology:
\begin{description}
\item[T] Turkic origin. This includes compound forms and affixed forms whose constituents are all Turkic. For example, the Turkmen for
`manager', \emph{\'{y}olba\c{s}\c{c}y}, is marked T because its compound base, \emph{\'{y}ol} with \emph{ba\c{s}}, and affix \emph{-\c{c}y} are all Turkic in origin.
\item[A] Arabic origin, to include words borrowed indirectly through another language such as Persian. For example, the word in every Turkic
language for `book' is marked A for all eight Turkic languages. Because variations on the Arabic form /kita:b/ exist in
every Turkic language, in Persian, and in other languages of the Islamic world, it is difficult to tease out the word's trajectory into a language such as Kyrgyz. The burden of researching these fine distinctions is not placed on the annotator, as explained below.
\item[P] Persian origin, not including Arabic words possibly borrowed through Persian. An example is the word for `color' in many Turkic
languages, from the Persian /r\ae ng/.
\item[R] borrowed from Russian, including words that are ultimately of French origin. 
\item[F] French origin, not including ultimately French words borrowed from Russian. Direct French loans occur almost exclusively in Turkish. An
example is the word for `station' in Turkish, \emph{istasyon}.
\item[E] English origin. For example the word for `basketball' in every language.
\item[I] Italian origin. Usually of importance only to specific domains in Turkish.  
\item[G] Greek origin. For example, the word in Azerbaijani, Turkish, Turkmen, Uyghur, and Uzbek for `box' comes from the Greek
$\kappa o\upsilon\tau\imath$. 
\item[C] Chinese origin, usually Mandarin and usually of importance only to Uyghur. An example is the word for `mushroom' in Uyghur,
(\emph{mogu}).
\item[Q] unknown or inconclusive origin.
\end{description}

The careful reader will have noticed that there is an inconsistency in that words of ultimately Arabic origin borrowed through
Persian are marked as A, but words of ultimately French origin borrowed through Russian are marked as R. There are two reasons for this. 
The first is annotator efficiency. Making the judgment that a word is ultimately of Arabic origin is much easier than having to figure
out whether it was borrowed from Arabic or indirectly from Persian. For the Russian/French situation, the distinction is much easier to 
make.
To begin with, the Russian loanwords occur almost exclusively in former USSR languages and the French loanwords occur almost exclusively in
Turkish. Also, the orthography often gives clear cues for making this distinction, as Russian loanwords consistently retain 
characteristically Russian letters.

\subsubsection{Multi-Language Exceptions}

We also define other codes that categorize certain complex words that do not fall into any of the categories described in 
subsection~\ref{etymology}. Other etymological annotation studies, such as the Loanword Typology project and its 
World Loanword Database \cite{LWT2009}, have instructed linguists to pass over such complex words and optionally flag them
as ``contains a borrowed base,'' etc.
Our annotation system requires that these words, which are very common in Turkic languages, be 
annotated according to more fine grained categories.

The following are our multi-language exception codes:
\begin{description}
\item[X] Compound words where the constituents are from different origins. For example, the Tatar word for `truck', (\emph{y\"{o}k mashinas\i}), is to be
marked X since it contains Russian-origin (\emph{mashina}), 'machine, vehicle' in compound with Tatar (\emph{y\"{o}k}), `baggage,cargo.'  In contrast, the Turkish compound
word for thunder, \emph{g\"{o}k g\"{u}rlemesi}, will be marked T because all of its constituents are Turkish.
\vspace{-.1cm}
\item[V] A verb formed by combining a non-Turkic base with a Turkic auxiliary verb or denominal affix. For example, the verb `to repeat' in
Azerbaijani, Tatar, and Turkish, because it consists of a noun borrowed from the Arabic /takra:r/ plus a Turkic auxiliary verb \emph{et-} or
\emph{it-}. 
\vspace{-.1cm}
\item[N] A nominal consisting of a non-Turkic base bearing one or more Turkic affixes, in cases where removing the affixes
results in a form that can plausibly be found elsewhere in the data or in a loan language dictionary. For example, the Kazakh word for
`baker,' (\emph{nawbaysh\i}), is composed of a Persian-origin base, from /na:nva:/, `baker', and a suffix that indicates a person 
associated with a
profession, (\emph{-sh\i}). The Turkmen word for `baker,' (\emph{\c{c}\"{o}rek\c{c}i}), on the other hand, will be marked T, because 
both its base
(\emph{\c{c}\"{o}rek}) and affix (\emph{-\c{c}i}) are Turkic. 
\end{description} 

Table~\ref{t:cognatesAndEtymologyExample} shows an example of an entry that has been fully annotated for both cognates and etymology. 

\begin{table*}[hp] 
\begin{center}
\begin{tabular}{|l|l|l|l|l|l|l|l|} 
      \hline
      Azerbaijani & Kazakh & Kyrgyz & Tatar & Turkish & Turkmen & Uyghur & Uzbek \\ \hline
      stul & (or\i nd\i q) & (orunduk) & (ur\i nd\i k) & sandalye & stul & (orunduq) & kursi \\ \hline
      1R & 2T & 2T & 2T & 3A & 1R & 2T & 4A \\ \hline
\end{tabular}
\caption{Example with complete annotation both for cognates and etymology. The English gloss here is `chair.'}
\label{t:cognatesAndEtymologyExample}
 \end{center}
\end{table*}

\section{Inter-Annotator Agreement}

We pilot-tested our annotation system with two annotators on 400 
etymology annotations.\footnote{Table~\ref{t:annotationAgreement} has 392 entries because the annotators claimed 
eight entries had multiple translations for the same English gloss.} 
Both annotators have studied linguistics. Also, both are native English speakers with experience studying or speaking multiple 
Turkic languages, Persian, and Arabic. Training consisted of studying the authors' annotation manual and asking any follow-up questions. 
Both annotators made approximately 240 annotations per hour. 

Table~\ref{t:annotationAgreement} shows the contingency matrix for annotating
the 400 entries.\footnote{We left out
columns for English, Greek, Italian, and Chinese, which were not relevant for the 50 entries (according to unanimous agreement of our annotators).}
From Table~\ref{t:annotationAgreement} it is immediate that agreement is substantial, and when there is disagreement it is largely for the difficult cases of 
inconclusive origin and the multi-language exceptions: Q, X, V, and N.
We measured inter-annotator agreement using Cohen's Kappa \cite{cohen1960} and found Kappa = 0.5927 (95\% CI = 0.5192 to 0.6662). 
If we restrict attention to only the instances where neither of the annotators marked an inconclusive origin or multi-language exception, then Kappa is 0.9216,
generally considered high agreement. This shows that our annotation system is feasible for use and also shows that to improve the system we might focus efforts on
finding ways to increase agreement on the annotation of the exceptional cases (Q, X, V, and N).  

\begin{table}[hp] 
\begin{center}
\begin{tabular}{|l|l|l|l|l|l|l|l|l|l|} 
      \hline
        & T   & A & P & R & F & Q & X  & V & N \\ \hline
      T & 160 & 8 & 2 & 0 & 0 & 3 & 10 & 6 & 1 \\ \hline
      A & 0 & 56 & 2 & 6 & 0 & 1 & 0 & 1 & 0 \\ \hline
      P & 0 & 0 & 31 & 0 & 0 & 0 & 1 & 0 & 0 \\ \hline
      R & 0 & 0 & 0 & 32 & 1 & 0 & 0 & 0 & 0 \\ \hline
      F & 0 & 0 & 0 & 0 &  5 & 0 & 0 & 0 & 0 \\ \hline
      Q & 12 & 5 & 0 & 2 & 0 & 0 & 2 & 3 & 0 \\ \hline
      X & 2 & 0 & 1 & 5 & 0 & 0 & 17 & 8 & 0 \\ \hline
      V & 0 & 1 & 0 & 0 & 0 & 0 & 0 & 0 & 0 \\ \hline
      N & 0 & 0 & 1 & 0 & 0 & 0 & 6 & 0 & 1 \\ \hline      
\end{tabular}
\caption{Table of Counts for two annotators' etymological conjectures on 392 words. 
Annotator 1's conjectures follow the horizontal axis, and annotator 2's
the vertical.}
\label{t:annotationAgreement}
\vspace{-.6cm}
 \end{center}
\end{table}

\section{Conclusions and Future Work}

The Turkic languages are a promising candidate family of languages to benefit from automated translation lexicon induction. A necessary step in that
direction is the creation of annotated data for cognates and etymology. However, this annotation is not straightforward, as the Turkic
languages exhibit extensive and diverse etymological relationships among words. Some distinctions are difficult for annotators to make and some are
easier. Also, some distinctions are expected to be more useful than others for automating cross-lingual applications among the Turkic languages.
We presented an annotation methodology that balances the research effort required of the annotator with the expected
value of the annotations. We surveyed and explained the wide range of the most important relationships observed in the Turkic languages
and how to annotate them. When we finish the annotations, we would like to make the annotated data available as long as it is legal under
copyright laws for us to do so. 
Finally, we hope that our annotation
system and the associated discussion can be useful for other teams that are annotating Turkic resources, and perhaps parts of it can be
useful for annotating resources for other language families as well.

\bibliographystyle{lrec2006}
\bibliography{paper} 

\begin{thebibliography}{}

\bibitem[\protect\citename{Cohen}1960]{cohen1960}
J.~Cohen.
\newblock 1960.
\newblock A coefficient of agreement for nominal scales.
\newblock {\em Educational and Psychological Measurement}, 20:37--46.

\bibitem[\protect\citename{Haspelmath and Tadmor}2009]{LWT2009}
Martin Haspelmath and Uri Tadmor.
\newblock 2009.
\newblock The {L}oanword {T}ypology project and the {W}orld {L}oanword
  {D}atabase.
\newblock In Martin Haspelmath and Uri Tadmor, editors, {\em Loanwords in the
  World's Languages: A Comparative Handbook}, pages 1--34, Berlin. Walter de
  Gruyter.

\bibitem[\protect\citename{Kirchner}2006]{kirchner2006}
Mark Kirchner.
\newblock 2006.
\newblock Kirghiz.
\newblock In Lars Johanson and \'{E}va \'{A}.~Csat\'{o}, editors, {\em The
  Turkic Languages}, pages 344--356, New York. Routledge.

\bibitem[\protect\citename{Kondrak}2001]{kondrak2001}
Grzegorz Kondrak.
\newblock 2001.
\newblock Identifying cognates by phonetic and semantic similarity.
\newblock In {\em Proceedings of the second meeting of the North American
  Chapter of the Association for Computational Linguistics on Language
  technologies}, NAACL '01, pages 1--8, Stroudsburg, PA, USA. Association for
  Computational Linguistics.

\bibitem[\protect\citename{Mann and Yarowsky}2001]{mann2001}
Gideon~S. Mann and David Yarowsky.
\newblock 2001.
\newblock Multipath translation lexicon induction via bridge languages.
\newblock In {\em Proceedings of the second meeting of the North American
  Chapter of the Association for Computational Linguistics on Language
  technologies}, NAACL '01, pages 1--8, Stroudsburg, PA, USA. Association for
  Computational Linguistics.

\bibitem[\protect\citename{\"{O}ztop\c{c}u \bgroup et al.\egroup
  }1996]{oztopcu1996}
Kurtulu\c{s} \"{O}ztop\c{c}u, Zhoumagaly Abuov, Nasir Kambarov, and Youssef
  Azemoun.
\newblock 1996.
\newblock {\em Dictionary of the Turkic Languages}.
\newblock Routledge, New York.

\bibitem[\protect\citename{R\'{o}na-Tas}2006]{ronatas2006}
Andr\'{a}s R\'{o}na-Tas.
\newblock 2006.
\newblock The reconstruction of {P}roto-{T}urkic and the genetic question.
\newblock In Lars Johanson and \'{E}va \'{A}.~Csat\'{o}, editors, {\em The
  Turkic Languages}, pages 67--80, New York. Routledge.

\bibitem[\protect\citename{Schafer and Yarowsky}2002]{schafer2002}
Charles Schafer and David Yarowsky.
\newblock 2002.
\newblock Inducing translation lexicons via diverse similarity measures and
  bridge languages.
\newblock In {\em proceedings of the 6th conference on Natural language
  learning - Volume 20}, COLING-02, pages 1--7, Stroudsburg, PA, USA.
  Association for Computational Linguistics.

\bibitem[\protect\citename{{T\"{u}rk Dil Kurumu}}1942]{TDK1942}
{T\"{u}rk Dil Kurumu}.
\newblock 1942.
\newblock {\em Felsefe ve Gramer Terimleri}.
\newblock Cumhuriyet Bas\i mevi, Istanbul.

\bibitem[\protect\citename{van~der Ark \bgroup et al.\egroup
  }2007]{vanderArk2007}
Ren\'{e} van~der Ark, Philippe Mennecier, John Nerbonne, and Franz Manni.
\newblock 2007.
\newblock Preliminary identification of language groups and loan words in
  {C}entral {A}sia.
\newblock In {\em Proceedings of the RANLP Workshop on Computational
  Phonology}, pages 12--20, Borovetz, Bulgaria.

\end{thebibliography}

\end{document}